\documentclass[sigconf]{acmart}
\renewcommand\footnotetextcopyrightpermission[1]{} 

\setcopyright{none}
\copyrightyear{}
\acmYear{}
\acmDOI{}
\acmConference[Document Intelligence Workshop at KDD]{The Second Document Intelligence Workshop at KDD}{2021}{Virtual Event}
\acmPrice{}
\acmISBN{}

\usepackage{hyperref}
\usepackage{latexsym}

\usepackage[utf8]{inputenc}

\usepackage{url}
\usepackage{xcolor}

\usepackage{graphicx} 
\usepackage{amsmath} 

\newcommand{\masktoken}{\texttt{[MASK]}}

\begin{document}

\title{Position Masking for Improved Layout-Aware Document Understanding}

\author{Anik Saha}
\affiliation{%
    \institution{Rensselaer Polytechnic Institute}
    \streetaddress{110 8th St}
    \city{Troy, NY}
    \country{USA}
}
\email{sahaa@rpi.edu}

\author{Catherine Finegan-Dollak}
\affiliation{%
    \institution{IBM Research}
    \streetaddress{1101 Kitchawan Rd}
    \city{Yorktown Heights, NY}
    \country{USA}
}
\email{cfd@ibm.com}

\author{Ashish Verma}
\affiliation{%
    \institution{IBM Research}
    \streetaddress{1101 Kitchawan Rd}
    \city{Yorktown Heights, NY}
    \country{USA}
}
\email{ashish.verma1@ibm.com}

\begin{abstract}
Natural language processing for document scans and PDFs has the potential to enormously improve the efficiency of business processes. 
Layout-aware word embeddings such as LayoutLM \cite{xu2020layoutlm-KDD} have shown promise for classification of and information extraction from such documents.
This paper proposes a new pre-training task called \textit{position masking} that can improve performance of layout-aware word embeddings that incorporate 2-D position embeddings. 
We compare models pre-trained with only language masking against models pre-trained with both language masking and position masking, and we find that position masking improves performance by over 5\% on 
a form understanding task.
\end{abstract}




\maketitle

\section{Introduction}\label{sec:intro}
\begin{figure}[t]
    \centering
    \includegraphics[width=0.8\columnwidth]{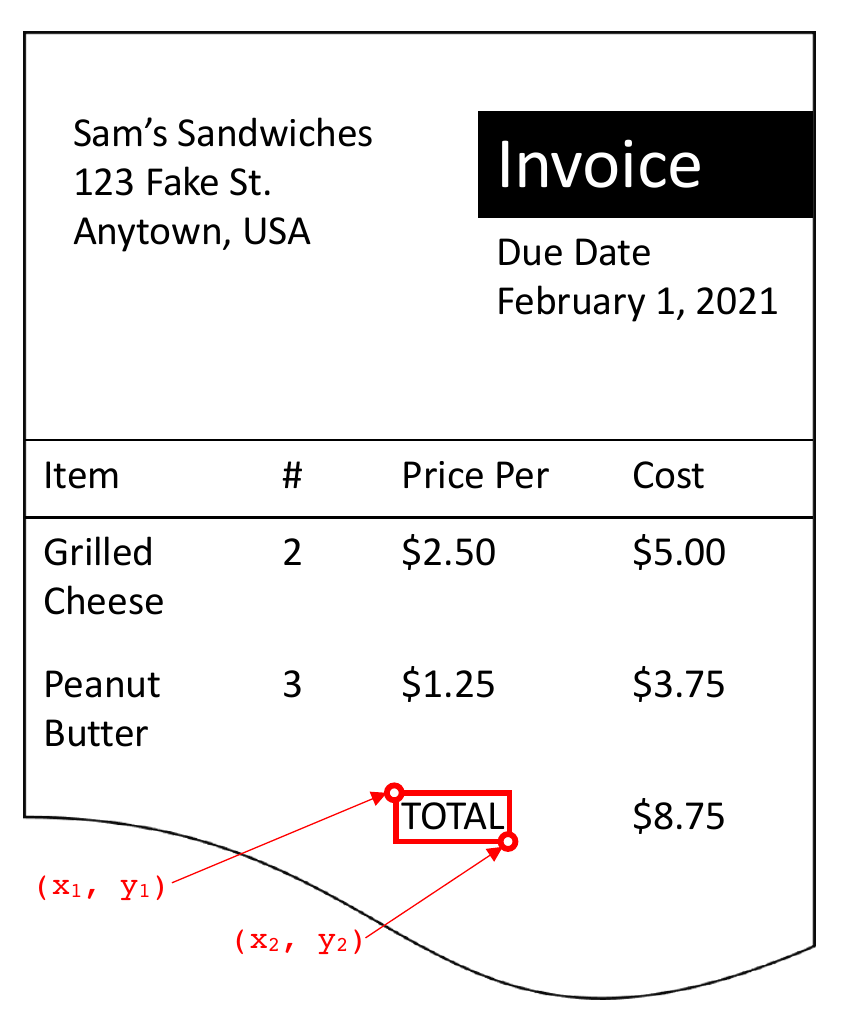}
    \caption{An example invoice, with one bounding box and coordinates shown. A human can use the positions of text to help determine the total cost.}
    \label{fig:invoice-example}
\end{figure}

For many real-world documents, layout---how the text is positioned on a two-dimensional page---is essential to meaning.
For example, the layout of addresses on an envelop tells us who is the sender and who the receiver.
Similarly, in the invoice in Figure \ref{fig:invoice-example}, a human can easily tell from the layout that the total due is \$8.75.
Recent systems that combine information about document layout with natural language processing techniques have shown improved performance on tasks such as 
information extraction \cite{katti-etal-2018-chargrid,Denk2019,xu2020layoutlm-KDD,lockard-etal-2020-zeroshotceres}.
Such techniques hold great promise for many business processes that must be performed on scanned or native PDF formatted document images.

Information extraction (IE) from forms is essential to many business processes, from identifying the amount due on an invoice to identifying the shipping address on a purchase order to getting the name of a borrower on a loan application. 
Text-based IE approaches \cite{stanovsky-etal-2018-supervised} that are successful in other domains, such as news articles and Wikipedia, 
don’t take advantage of the essential nature of forms: keys and their values are often spatially grouped.
Systems such as LayoutLM \cite{xu2020layoutlm-KDD} that leverage this information have an advantage over text-only systems like BERT \cite{devlin2018bert}. 

This paper's contribution is a new pre-training task for layout-aware text embeddings: position masking.
A masked language model (MLM) pre-training task, like that used by LayoutLM \cite{xu2020layoutlm-KDD}, masks token embeddings and tries to predict the correct token based on context---the unmasked components of the token, such as its position, and the other tokens in the input sequence.
Our model additionally masks 2-D position embeddings and attempts to predict the correct position based on context.
We show that adding the position-masking pre-training task to LayoutLM results in an absolute improvement of 5\% on a form-understanding task.
The proposed pre-training task would be compatible with any text embeddings that incorporate 2-D position embeddings, such as the new LayoutLMv2 \cite{Xu2020}.

\section{Related Work}\label{sec:related}
Several approaches have recently emerged for layout-aware document representations, particularly aimed at information extraction (IE).
\cite{holt-chisholm-2018-extracting} engineered a token representation that included word shape and positional information.
\cite{Majumder2020}'s token representation incorporated a 2-D position embedding for both the token itself and a neighborhood of nearby tokens.
\cite{wang-etal-2020-docstruct} sought to extract a hierarchical structure among text fragments on form pages using a feature fusion model to combine text, layout, and visual features.
\cite{liu-etal-2019-graph} and \cite{lockard-etal-2020-zeroshotceres} both constructed document graphs where nodes were text boxes and edge embeddings incorporated spatial information.

Some works transform a 2-D page into a 3-D tensor.
They assign each pixel on a page a vector
, generating a $vector size \times height \times width$ tensor.
This representation is then passed to a CNN encoder-decoder model for closed-class IE.
In Chargrid \cite{katti-etal-2018-chargrid}, each character is assigned a one-hot encoding.
BERTgrid \cite{Denk2019} replaces the one-hot character vectors with BERT embeddings for tokens.
\cite{Kerroumi2020} incorporated the RGB values for pixels into Chargrid.
A downside to these techniques is the size of the tensors that result, making scalability a problem.


LayoutLM \cite{xu2020layoutlm-KDD} is a transformer-based MLM that adds 2-D position embeddings to the input text embeddings for a BERT-like model.
Our work is complementary to LayoutLM, adding a new pre-training task to enhance its performance.
In work concurrent with ours, \cite{Xu2020} improved on LayoutLM by using a multi-modal transformer and adding new pre-training tasks.
Our technique could also be combined with theirs.


\cite{wang-chen-2020-position} compares learned position embeddings from BERT, RoBERTa, and GPT-2, with sinusoidal position embeddings.
Its focus was on 1-D position embeddings, though; to our knowledge, no similar study has been conducted on 2-D position embeddings.

\section{Approach}\label{sec:model}
\begin{figure*}[ht]
    \centering
    \includegraphics[width=0.9\textwidth]{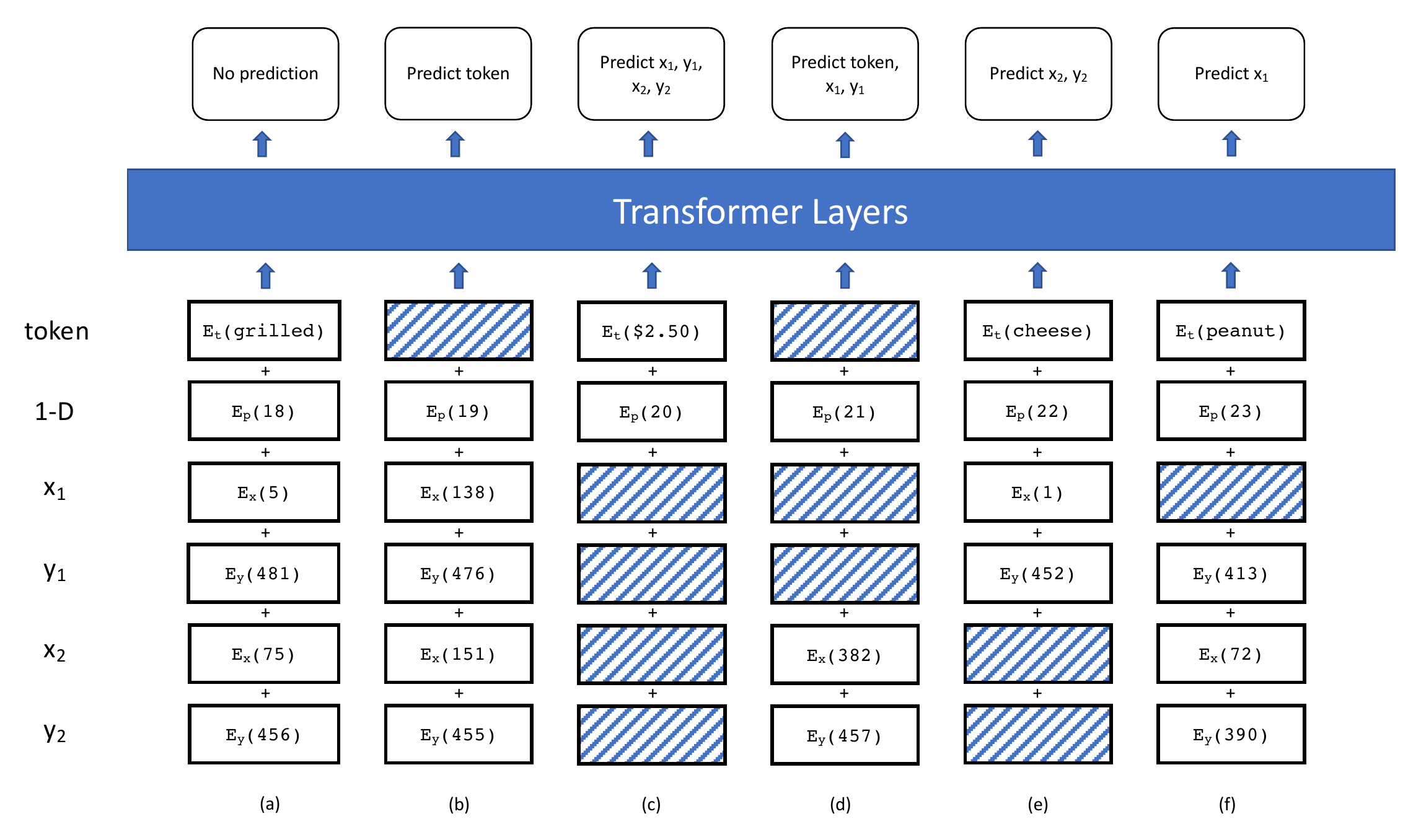}
    \caption{Position masking and language masking applied to a subset of tokens from the invoice in Figure \ref{fig:invoice-example}. Shaded boxes indicate embeddings that have been masked.
    Segment embeddings omitted for brevity. 
    Column a: No masking applied. Column b: Token masking only; can be used in BERT, LayoutLM, or our model. Column c: Full position masking. Column d: Partial position masking of $x_1$ and $y_1$ and token masking. Column e: Partial position masking of $x_2$ and $y_2$. Column f: Partial position masking of $x_1$ only.} 
    \label{fig:model_diagram}
\end{figure*}

Our approach, illustrated in Figure \ref{fig:model_diagram}, builds on LayoutLM \cite{xu2020layoutlm-KDD}, 
which we briefly review before describing the proposed improvement.

\subsection{Background: LayoutLM}
LayoutLM passes input text embeddings through transformer layers. 
Each of LayoutLM's input text embeddings is the sum of a token embedding, a 1-D position embedding, a segment embedding,\footnote{BERT uses segment IDs when the input text includes two parts; \textit{e.g.}, for detecting semantic similarity between two sentences, the segment marks which tokens are part of which sentence. Here, all segment IDs are 0.} and 2-D position embeddings.

More formally, suppose we have a vocabulary $V$ and a document image $D$ comprised of tokens $t_0, t_1...t_N$ ($t_i \in V$) with bounding boxes $b_0, b_1...b_N$ and segments $s_0, s_1...s_N$.
Token $t_i$'s 1-D position is $i$---that is, its position in the input sequence.
Bounding box $b_i$ is defined by its top left corner at $(x^i_1, y^i_1)$ and its bottom right corner at $(x^i_2, y^i_2)$, as illustrated in Figure \ref{fig:invoice-example}. Its width is $w_i=x^i_2-x^i_1$ and height is $h_i=y^i_2-y^i_1$.
Let $\mathrm{E}(x)$ be an embedding function, with the subscript 
$t$ for token embeddings, $p$ for 1-D position embeddings, $s$ for sequence embeddings, $x$ for $x$-coordinate embeddings, $y$ for $y$-coordinate embeddings, $w$ for width embeddings, and $h$ for height embeddings. The input text embedding for the $i$-th token is
\begin{equation} \label{eq:embedding-sum}
    \begin{split}
              E_i = \mathrm{E}_{t}(t_i) + \mathrm{E}_{p}(i) + \mathrm{E}_{s}(s_i) 
                     + \mathrm{E}_{x}(x^i_1)
                     + \mathrm{E}_{y}(y^i_1) \\
                     + \mathrm{E}_{x}(x^i_2) 
                     + \mathrm{E}_{y}(y^i_2) 
                     + \mathrm{E}_{w}(w_i)
                     + \mathrm{E}_{h}(h_i)  
    \end{split}
\end{equation}

LayoutLM uses a masked language model (MLM) pre-training task.
In the MLM task, 
for some randomly selected subset of input tokens $J$, for each $j \in J$, $t_j$ is replaced by \masktoken, so the first term of Equation \ref{eq:embedding-sum} becomes $\mathrm{E}_{t}(\masktoken)$.
The token is the only component of the input text embedding that is masked, as illustrated in Figure \ref{fig:model_diagram}(b).
The transformer model is trained to use context---all of the unmasked components of the embedding $E_j$ and the surrounding embeddings ($E_i$ for all $i \neq j$)---to predict the missing token, $\hat{t}_j \in V$. 
MLM loss is 
\begin{equation}
\mathcal{L}_{MLM} = \sum_{j\in J} \mathrm{Cross Entropy}(t_j, \hat{t}_j)
\end{equation}

\subsection{Position Masking}

This work introduces \textit{position masking}, as shown in Figure \ref{fig:model_diagram}.
We replace a fraction of \textit{positions} with the coordinate assigned to \masktoken\ (the maximum x and y coordinate in the 2-D matrix) and train the model to predict what the true position was.
Similar to the MLM task, the model bases its predictions on the context.
Predicting the 2-D position of a token on a page will, we believe, force the model to better learn the relationship of layout and text.

We randomly choose a set $K$ of tokens for position masking.
Since $J$ (the set of masked \textit{tokens}) and $K$ are selected randomly and independently, a small fraction may overlap, as in Figure \ref{fig:model_diagram}(d).
In these cases, the model is forced to rely more heavily on the surrounding text for context, rather than on unmasked components of the same input text embedding.
This encourages the model to learn inter-token relationships of layout and text.

Since bounding boxes consist of multiple elements, position masking may be full or partial.
Full position masking (Figure \ref{fig:model_diagram}(c)) selects a fraction of tokens and replaces all the coordinates with their \masktoken\ values.
Partial position masking (Figure \ref{fig:model_diagram}(d)-(f)) replaces only some coordinates.

Position masking can be framed as either regression or classification. 
As a classification problem, the labels are the integers from $[0, m]$, where $m$ is the maximum value of a pixel on the axis being predicted. 
Classification has the advantage of more closely mirroring MLM.
However, regression may be more logical where we are trying to predict a measurable value rather than a categorical one.
We define $g(x, \hat{x})$ to be a cross entropy loss function when we use a classifier and a smooth-L1 loss \cite{Girshick2015} when we use regression. 
If we mask only $x_1$, then position-masking loss is
\begin{equation}
\mathcal{L}_{PM} = \sum_{k\in K} g(x^1_k, \hat{x}^1_k)
\end{equation}
When we do full position masking, we average the equivalent losses for all position embeddings. 

We train the MLM and masked position model simultaneously, using the loss function
\begin{equation}
\mathcal{L} = \mathcal{L}_{MLM} + \lambda \mathcal{L}_{PM}
\end{equation}
where $\lambda$ is a hyperparameter to weight the position masking loss.

\section{Experiments}
\subsection{Systems}
We compare LayoutLM with no position masking against LayoutLM with several varieties of position masking.
We used the publicly available implementation of  LayoutLM.\footnote{\url{https://github.com/microsoft/unilm/tree/master/layoutlm}. We implemented the pre-training script ourselves since the public release did not include it.} 
All experiments use the BASE-sized model with the Masked Visual-Language Model (MVLM) loss but not Multi-label Document Classification (MDC) loss function.

Our position masking variations are as follows:
PosMaskx1 uses partial position masking of the $x_1$ embedding, while PosMaskFull masks all four coordinate embeddings.
CELoss trains a classifier and RegLoss a regressor.

Position masking carries a risk that the model will learn the relationship of coordinates to height and width, rather than of positions to tokens.
We therefore excluded the height and width embeddings from the PosMask conditions.


\subsection{Pre-training}
We pre-train all models on 500,000 document images from tobacco litigation settlement documents.\footnote{\label{note:data}The original LayoutLM paper pre-trained on tobacco litigation documents from IIT-CDIP \cite{Lewis2006} 
IIT-CDIP was unavailable to the public during much of 2020-21 (See discussions such as
\url{https://github.com/microsoft/unilm/issues/250}), so we collected comparable documents.
Moreover, in light of 
concerns about the environmental impact of enormous experiments \cite{strubell-etal-2019-energy}, 
and since the aim of the current work is to compare models with and without position masking, rather than to exceed SOTA, we do not use 11 million pages, as LayoutLM's largest models did. 
Our pre-training size of 500,000 pages was the smallest pre-training size reported in \cite{xu2020layoutlm-KDD}, which outperformed the corresponding BERT model.}
Three hundred twenty thousand of these are from the train set of RVL-CDIP\footnote{\url{https://www.cs.cmu.edu/~aharley/rvl-cdip/}} \cite{Harley2015}, with labels removed.
We collected the remaining 180,000 from the University of California San Francisco Library's Truth Tobacco Industry Documents.\footnote{\url{https://www.industrydocuments.ucsf.edu/tobacco/}
Details of our collection method are in Appendix \ref{app:data}.
The IDs and page numbers for document images included in our dataset are available at \url{https://github.com/aniksh/tobacco_documents}.}
We use Tesseract OCR v.4.1\footnote{\url{https://github.com/tesseract-ocr/tesseract}} to obtain text and bounding boxes for all scanned document images.
Additional pre-training settings are in Appendix \ref{app:settings}.

\subsection{Evaluation}
We compare the performance of the pre-trained systems on the downstream task of form understanding.
Form understanding aims to identify keys (sometimes called questions) and values (sometimes called answers) on forms.
For instance, in Figure \ref{fig:invoice-example}, ``Due Date'' is a key with the value ``February 1, 2021.''
We evaluate on the FUNSD dataset\footnote{\url{https://guillaumejaume.github.io/FUNSD/}} \cite{jaume2019}.
It includes 199 (149 train/50 test) scans of forms, with gold labels for keys, values, headers, and ``other'' entity types. 

We fine tune for 100 epochs on the train set, then predict an entity tag for each token in the test set.
We report precision, recall, and $F_1$ for entity identification.
Following \cite{xu2020layoutlm-KDD}, we leave linking for future work.
We report the mean over 5 fine-tuning runs for each system.
We use a one-way ANOVA and a Tukey test to check statistical significance.


\begin{table}[ht]
    \setlength{\tabcolsep}{4pt}
    \centering
    \small
    \begin{tabular}{|lccc|} 
        \hline
         Model & Precision & Recall & F1 \\ \hline
         LayoutLM & 61.1 (0.8) & 69.2 (0.2) & 64.9 (0.5)\\
         \hline
         \multicolumn{4}{|l|}{PosMaskx1}\\
         \hspace{1em} with CE Loss & 66.8 (0.6) & \textbf{74.0} (0.5) & \textbf{70.2} (0.4)\\
         \hspace{1em} with Reg Loss & 65.5 (1.0) & 73.6 (0.6) & 69.3 (0.8)\\
         \multicolumn{4}{|l|}{PosMaskFull} \\
         \hspace{1em} with CE Loss & \textbf{67.1} (1.0) & 73.6 (0.6) & \textbf{70.2} (0.8)\\
         \hspace{1em} with Reg Loss & 64.9 (1.0) & 72.9 (1.0) & 68.7 (0.9)\\
         \hline
    \end{tabular}
    \caption{Mean (standard deviation) Precision, Recall, and F1 scores on FUNSD. PosMaskx1: partial position masking of the x1 coordinate. PosMaskFull: full position masking. CE Loss: cross entropy. Reg Loss: regression loss. Results are not directly comparable with \cite{xu2020layoutlm-KDD}'s Table 1 due to difference in training data.\textsuperscript{\ref{note:data}}}
    \label{tab:funsd}
\end{table}

\section{Results \& Discussion}\label{sec:results}
  

As shown in Table \ref{tab:funsd}, all of the position-masking models outperformed the baseline.
The differences are significant ($p<0.001$).
Two position-masking models outperformed LayoutLM without position masking by over 5\%.
Full position masking with regression loss was significantly worse than either of the cross entropy conditions ($p<0.05$). No other differences were significant.

Since our position masking implementation did not include height or width embeddings, we performed an ablation to determine whether masking the positions or removing height and width was responsible for the difference.
LayoutLM with no height or width embeddings and with no position masking achieved mean $F_1$ of 67.5, which is significantly better than the baseline ($p<0.001$), although the improvement was not as large the improvement from position masking.
The model with no height or width embeddings showed nearly the same improvement in recall as the position masking models (mean recall 72.7), but a much smaller increase in precision (mean precision 63.0).
Both position masking systems with cross entropy loss had significantly better $F_1$ scores than the ablated model ($p<0.001$), as did masking x1 with regression loss ($p<0.05$).

  


\section{Conclusion}\label{sec:conclusion}
This work has introduced a new pre-training task, position masking, to improve layout-aware text embeddings.
We have shown that adding position masking to a LayoutLM baseline model improved performance on a form understanding task. 
Future work should explore how well this technique can generalize to other tasks.

\bibliographystyle{ACM-Reference-Format}
\bibliography{key-value-sources.bib}

\appendix
\section{Supplemental Tobacco Documents Dataset}\label{app:data}
Our pre-training data incorporates 180,000 document images we collected from the University of California San Francisco's Truth Tobacco Industry Documents database 
(formerly known as the Legacy Tobacco Documents Library), the same source that \cite{Lewis2006} used to build IIT-CDIP originally.

We collected approximately 99,000 PDFs from the tobacco litigation documents using the search terms `(availability:public AND industry:tobacco)'.
We iterated through the files in an arbitrary order, dividing them into their individual pages, until we had over 300,000 pages.
After performing OCR on these pages using Tesseract, we manually inspected samples of the smallest hOCR output files. 
We determined that files under 1.8 kb in size typically did not include text except for whitespace, so we deleted pages that yielded such small output. 
From the remaining pages, we randomly sampled 180,000 to be our supplemental dataset.

To enable replication, we will provide a list of the specific pages in the supplemental dataset.

\section{Detailed Experimental Settings}\label{app:settings}
\paragraph{Pre-training.} We initialized the model parameters with the BERT-base language model. 
Distributed training was set up with 8 Tesla V100 GPUs, each with 32GB memory.
Batch size is 25 per GPU, so the total batch size is 200.
We use the same AdamW optimizer as LayoutLM with a learning rate of 5e-5 without weight decay.
A linear learning rate schedule was used that goes from the initial value to zero at the end of all epochs.
The gradient norm was clipped at 1.
For position masking loss, the weight $\lambda$ is set to 1 based on initial experiments on a small pre-training dataset.

\paragraph{Fine Tuning}
Fine tuning was done in a single GPU as the dataset is small.
So the batch size is 25.
Same optimizer and learning rate scheduler as pre-training was used.

\end{document}